\documentclass[11pt]{article}
\usepackage{eacl2023}
\usepackage[T1]{fontenc}
\usepackage[utf8]{inputenc}
\usepackage{newtx/tex/newtx}
\usepackage{latexsym}
\usepackage{inconsolata}
\usepackage{multirow}
\usepackage{makecell}

\usepackage{graphicx}
\usepackage{pgf}
\usepackage{booktabs}
\usepackage{colortbl}
\usepackage[inline]{enumitem}
\usepackage{etoolbox}
\usepackage{tabularx}
\usepackage{floatrow}
\usepackage{microtype}
\usepackage{xfp}
\usepackage{twoopt}
\usepackage{xspace}
\usepackage{pifont}

\graphicspath{{./graphics/}}
\makeatletter\let\input@path\Ginput@path\makeatother

\newcommand{\miniscule}{\fontsize{5.25pt}{6.3pt}\selectfont}

\date{\today}

\newif\iffinalcopy%
\makeatletter
  \ifacl@finalcopy%
    \finalcopytrue%
  \else
    \finalcopyfalse%
  \fi
\makeatother

\setlist{wide, nosep, mode=unboxed}
\newlist{romenum}{enumerate}{2}
\newlist{romenum*}{enumerate*}{2}
\newlist{alphenum}{enumerate}{2}
\newlist{alphenum*}{enumerate*}{2}
\setlist[romenum]{label= (\roman*)}
\setlist[romenum*]{label= (\roman*)}
\setlist[alphenum]{label= (\alph*)}
\setlist[alphenum*]{label= (\alph*)}
\setenumerate{label= (\arabic*)}

\newcommand{\cmark}{{\color[HTML]{00715E}{\ding{51}}}}%
\newcommand{\xmark}{{\color[HTML]{961C26}{\ding{55}}}}%

\newcommand{\equal}[1]{\color[HTML]{AE8E00}{#1}}

\newcommand{\faint}[1]{\textcolor{gray}{#1}}
\newcommand{\xmoverscore}[1][]{\textsc{XMoverScore\ifstrempty{#1}{}{ (#1)}}}
\newcommand{\distilscore}{\textsc{DistilScore}}
\newcommandtwoopt{\metric}[2][][]{\textsc{UScore$^{\text{#2}}_{\text{#1}}$}\xspace}
\newcommand{\sentsim}[1][]{\textsc{SentSim\ifstrempty{#1}{}{ (#1)}}}
\newcommand{\prism}[1][]{\textsc{Prism\ifstrempty{#1}{}{-#1}}}
\newcommand{\bartscore}{\textsc{BARTScore}}
\newcommand{\bertscore}{\textsc{BERTScore}}
\newcommand{\bleu}{\textsc{BLEU}}
\newcommand{\rouge}{\textsc{ROUGE}}
\newcommand{\monotransquest}{\textsc{MonoTransQuest}}
\newcommand{\comet}{\textsc{COMET}}
\newcommand{\cometqe}{\textsc{COMET-QE}}
\newcommand{\bleurt}{\textsc{BLEURT}}
\newcommand{\unite}{\textsc{UniTE}}

\newcommand{\mbert}{\textsc{mBERT}}
\newcommand{\xlmr}{\textsc{XLM-R}}
\newcommand{\mbart}{\textsc{mBART}}
\newcommand{\bart}{\textsc{BART}}
\newcommand{\fastalign}{\textsc{fast-align}}
\newcommand{\awesomealign}{\textsc{awesome-align}}
\newcommand{\fasttext}{\textsc{fastText}}
\newcommand{\gpt}[1][2]{\textsc{GPT-#1}}
\newcommand{\type}[1]{\textsc{TYPE-#1}}
\newcommand{\wxling}{w_{\text{xlng}}}
\newcommand{\wpseudo}{w_{\text{pseudo}}}
\newcommand{\wlm}{w_{\text{lm}}}
\newcommand{\wwmd}{w_{\text{wrd}}}
\newcommand{\wcos}{w_{\text{snt}}}
\newcommand{\pearson}{Pearson's \textit{r}}

\makeatletter
\newcommand\scaleinput[1]{%
  \small
  \scalebox{\fpeval{\f@size/9.6}}{\input{#1}} %chktex 36
}
\newcommand\tightscaleinput[1]{%
  \miniscule%
  \scalebox{\fpeval{\f@size/9.6}}{\input{#1}} %chktex 36
}
\makeatother

\newcommand\pref[2][https://]{%
  \href{#1#2}{\nolinkurl{#2}}%
}

\newcommand\mail[2][mailto:]{%
  \href{#1#2}{\normalcolor\nolinkurl{#2}}%
}

\DeclareMathOperator{\wmd}{WMD}

\DeclareMathOperator{\lm}{LM}

\DeclareMathOperator{\embed}{E}
\DeclareMathOperator{\m}{\mathfrak{m}}
\DeclareMathOperator{\margin}{margin}

\title{\metric{}:
An Effective Approach to Fully \\ Unsupervised Evaluation Metrics for Machine Translation}

\author{%
  Jonas Belouadi\\
  \mail{jonas.belouadi@uni-bielefeld.de}
  \And%
  Steffen Eger\\
  \mail{steffen.eger@uni-bielefeld.de}
  \AND%
  {\rm Natural Language Learning Group (NLLG)}\\
  Faculty of Technology, Bielefeld University\\
  \pref{nl2g.github.io}
}

\begin{document}
\maketitle
\begin{abstract}
  The vast majority of evaluation metrics for machine translation are
  supervised, i.e.
  \begin{romenum*}[before=\unskip{, }]
    \item are trained on human scores,
    \item assume the existence of reference translations, or
    \item leverage parallel data.
  \end{romenum*}
  This hinders their applicability to cases where such supervision signals are
  not available. In this work, we develop \emph{fully unsupervised} evaluation
  metrics. To do so, we leverage similarities and synergies between evaluation
  metric induction, parallel corpus mining, and MT systems. In particular, we
  use an unsupervised evaluation metric to mine pseudo-parallel data, which we
  use to remap deficient underlying vector spaces (iteratively) and
  to induce an unsupervised MT system, which then provides pseudo-references as
  an additional component in the metric. Finally, we also induce unsupervised
  multilingual sentence embeddings from pseudo-parallel data. We show that our
  fully unsupervised metrics are effective, i.e., they beat supervised
  competitors on four out of five evaluation datasets. We make our code publicly
  available.\footnote{%
    \makeatletter%
      \iffinalcopy%
        \pref{github.com/potamides/unsupervised-metrics}%
      \else%
        Redacted for anonymity.%
      \fi%
    \makeatother%
  }
\end{abstract}

\section{Introduction}\label{sec:introduction}
Evaluation metrics are essential for judging progress in natural language
generation (NLG) tasks such as machine translation (MT) and summarization, as
they identify the state-of-the-art in a key NLP technology. Despite their wide
dissemination, classical lexical overlap evaluation metrics like
\bleu~\citep{papineni-2002} and \rouge~\citep{lin-2004} have difficulties
judging the quality of modern NLG
systems~\citep{mathur-etal-2020-tangled,marie-etal-2021-scientific},
necessitating novel   metrics that correlate better with humans. Lately, this
has been a very active research
area~\citep{Zhang2020BERTScoreET,zhao-2019,zhao2022discoscore,colombo-etal-2021-automatic,yuan2021bartscore}.\footnote{Of
course, the search for high quality metrics dates back at least to the
invention of \bleu{} and its predecessors.}

\begin{figure}[!tb]
  \centering
  \scriptsize
  \input{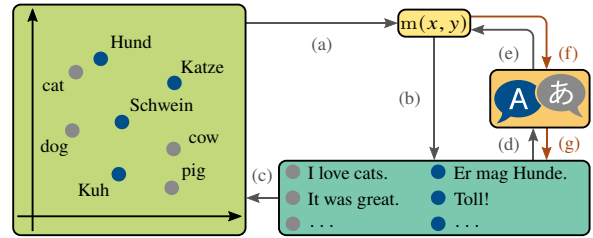}
  \caption{Relationship between metrics $\operatorname{m}$, vector spaces,
  parallel data, and MT systems: Metrics build on (potentially deficient)
  multilingual vector spaces~(a), and can be used to mine (pseudo-)parallel
  sentences~(b), which in turn can be used to improve deficient vector
  spaces~(c). (Pseudo-)parallel data can also be used to train MT systems~(d),
  which can generate pseudo-references~(e). Conversely, metrics can also be
  optimization criteria for MT systems, which in turn can generate additional
  pseudo-parallel data through translation~(f \& g; not explored in this
  work).}%
  \label{fig:self}
\end{figure}

Recently, more and more \emph{supervised} metrics are being proposed. E.g.,
\bleurt~\citep{sellam-2020} trains on human annotated datasets ranging from
5k-150k pairs, the \comet~\citep{rei-2020} models regress on 12k-370k data
points and \unite{}~\citep{wan-etal-2022-unite}, before fine-tuning on the same
data as \comet, pre-trains on 5m-10m parallel sentences. Of course, training on
larger amounts of data leads to better metrics {(measured on in-domain data)},
but also increases the risk of learning biases from the
data~\citep{poliak-etal-2018-hypothesis}---and limits the applicability to
domains and language pairs where supervision is available. Here, we go the
opposite route and try to minimize the amount of supervision as much as
possible.

We classify existing metrics making use of different types of supervision as
follows (cf. Table~\ref{tab:metrics-clf}).
\begin{description*}[itemjoin={{. }}, after={{.}}]
  \item[\type{1}] metrics are \emph{trained} on human assessments such as
    Direct-Assessment (DA) or Post-Editing (PE) scores, and compare
    system outputs to either human
    references~\citep[reference-based;][]{sellam-2020,rei-2020} or directly to
    source texts~\citep[reference-free;][]{ranasinghe-2021}
  \item[\type{2}] metrics, by comparison, do not use human assessments for
    training but still require human references, i.e., are untrained and
    reference-based~\citep{yuan2021bartscore,zhao-2019,Zhang2020BERTScoreET}
  \item[\textnormal{Finally,} \type{3}] metrics are untrained (unlike \type{1})
    and reference-free (unlike \type{2}), i.e., do not use supervision as in
    \type{1~\textnormal{or}~2}. However, to work well, they still rely on
    parallel data~\citep{Zhao2020OnTL,Song2021SentSimCS}, which is considered a
    form of supervision, e.g., in the MT
    community~\citep{artetxe2018unsupervised,lample-etal-2018-phrase}
\end{description*}
\begin{table}[t]
  \centering
  \begin{tabular}{l@{\hspace{5pt}}ccc}
    \toprule
    & \textbf{Training} & \textbf{References} & \textbf{Parallel Data}\\
    \midrule
    \textbf{\type{1}}     & \cmark\ & (\cmark) & (\cmark) \\
    \textbf{\type{2}}     & \xmark\ &  \cmark\ & (\cmark) \\
    \textbf{\type{3}}     & \xmark\ &  \xmark\ &  \cmark\ \\
    \midrule
    \textbf{Unsupervised} & \xmark\ &  \xmark\ &  \xmark\ \\
    \bottomrule
  \end{tabular}
  \caption{Different types of supervision used by \type{1/2/3} metrics compared
  to unsupervised metrics. Checkmarks surrounded by parentheses denote optional
  supervision signals.}%
  \label{tab:metrics-clf}
\end{table}

In contrast, we aim for \emph{fully unsupervised evaluation metrics} (for MT)
that do not use any form of supervision (cf. Table~\ref{tab:metrics-clf}). In
addition, subject to the constraint that no supervision is allowed, our metrics
should be of \emph{maximally high quality}, i.e., correlation with human
assessments. We have two use cases in mind:
\begin{alphenum*}
  \item Such \emph{sample efficiency}\footnote{We use the term sample
    efficiency in a generalized sense to denote the amount of supervision
    required.} is a prerequisite for the wide applicability of the metrics.
    This is especially important when we want to overcome the current
    English-centricity~\citep{anastasopoulos-neubig-2020-cross} of MT systems
    and evaluation metrics and also cover  low-resource \emph{languages} like
    Nepali or Sinhala~\citep{fomicheva-etal-2021-eval4nlp} and low-resource
    \emph{pairs} like Yoruba-German.\footnote{Neither Yoruba (a language spoken
    in Nigeria) nor German are classical low-resource languages. For German,
    this is clear and Yoruba is even included in \mbert{}, i.e., belongs to the
    languages with 100+ largest Wikipedias. Nonetheless, from own experience,
    we find it inherently difficult to obtain high-quality annotations for the
    language pair, as a result of few competent parallel speakers as well as
    technical difficulties (e.g., lack of adequate compute infrastructure in
    Nigeria).}
  \item Our fully unsupervised evaluation metrics should be considered
    \emph{strong lower bounds} for any future work that uses (mild) forms of
    supervision for metric induction, i.e., we want to push the lower bounds
    for newly developed \type{k} metrics.
\end{alphenum*}

To achieve our goals, we employ
\emph{self-learning}~\citep{He2020RevisitingSF,Wei2020TheoreticalAO} and in
particular, we leverage the following dualities to make our metrics maximally
effective, cf.\ Figure~\ref{fig:self}:
\begin{enumerate*}
    \item Evaluation metrics and NLG systems are closely related; e.g., a
      metric can be an optimization criterion for an NLG
      system~\citep{bohm-etal-2019-better}, and a system can conversely
      generate \emph{pseudo references} (a.o.) from which to improve a metric.
    \item Evaluation metrics and parallel corpus
      mining~\citep{Artetxe2019MarginbasedPC} are closely related; e.g., a
      metric can be used to mine parallel data, which in turn can be used to
      improve the metric~\citep{Zhao2020OnTL}, e.g., by remapping deficient
      embedding spaces.
\end{enumerate*}

Our contributions are:
\begin{romenum*}[font=\bfseries]
    \item We show that effective unsupervised evaluation metrics can be
      obtained by exploiting relationships with parallel corpus mining
      approaches and MT system induction;
    \item to do so, we explore ways to
      \begin{alphenum*}[itemjoin={{ and }}]
      \item make parallel corpus mining \emph{efficient} (e.g., overcome cubic
        runtime complexity)
      \item induce unsupervised multilingual sentence embeddings from
        pseudo-parallel data
      \end{alphenum*};
    \item we show that {pseudo-parallel}  data can rectify deficient vector
      spaces such as \mbert{};
    \item we show that our metrics beat three state-of-the-art supervised
      metrics on four of five datasets we evaluate on.
\end{romenum*}

\section{Background}
We take inspiration from three recent supervised (reference-free; \type{3})
metrics: \xmoverscore~\citep{Zhao2020OnTL},
\distilscore~\citep{Reimers2020MakingMS}, and
\sentsim~\citep{Song2021SentSimCS}. Below, we review key aspects of them, and
show where supervision plays a role.

\subsection{\xmoverscore}\label{sec:xmover}
Central to \xmoverscore{} is the use of Word Mover's Distance (WMD) as a
similarity  between two sentences~\citep{Zhao2020OnTL}. WMD and further
enhancements are discussed below.

\paragraph{WMD}\label{par:wmd}
WMD is a distance function that compares sentences at the token
level~\citep{Kusner2015FromWE}, by leveraging word embeddings which in
\xmoverscore{}'s case come from \mbert~\citep{devlin2018bert}. From a source
sentence $x$ and an MT hypothesis $y$, WMD constructs a  \emph{distance matrix}
$\mathbf{C}\in\mathbb{R}^{|x|,|y|}$, where $\mathbf{C}_{ij}$ is the distance
between two word embeddings, $\mathbf{C}_{ij}=\|\embed(x_i) - \embed(y_j)\|$;
$x_i,y_j$ index respective words in $x$, $y$. WMD uses this distance matrix to
compute the similarity of the two sentences. This  can be defined as the linear
programming problem
\begin{equation}\begin{split}
  \wmd(x,y)&= \min_\mathbf{F}
    \sum_{i=1}^{|x|}\sum_{j=1}^{|y|}\mathbf{F}_{ij}\mathbf{C}_{ij},
\end{split}\end{equation}
where $\mathbf{F}\in\mathbb{R}^{|x|,|y|}$ is an \emph{alignment matrix} with
$\mathbf{F}_{ij}$ denoting how much of word $x_i$ travels to word $y_j$.
Additional constraints prevent it from becoming a zero matrix.

\paragraph{Vector space remapping}
\citet{Zhao2020OnTL}, akin to similar earlier and subsequent work
\citep{Cao2020Multilingual,schuster-etal-2019-cross}, argue that the
monolingual subspaces of \mbert{} are not well aligned. As a remedy, they
investigate linear projection methods which post-hoc improve cross-lingual
alignments. We refer to this approach as vector space \emph{remapping}.
\xmoverscore{} explores two different remapping approaches, CLP and UMD\@. They
both leverage parallel data on sentence-level from which they extract
word-level alignments using \fastalign{}, which are then used for remapping. We
give more details in the Appendix~\ref{sec:remapping}.

\paragraph{Language Model}
\xmoverscore{} linearly combines WMD with the perplexity of a \gpt{} language
model~\citep{Radford2019LanguageMA}. Allegedly this penalizes ungrammatical
translations. This updates \xmoverscore{}s scoring function to
\begin{equation}\label{eq:lm-combine}
  \begin{split}
    \m(x,y) ={}& \wxling\wmd(x,y)+ \wlm\lm(y).
  \end{split}
\end{equation}
Here, $\wxling$, $\wlm$ are weights for the cross-lingual WMD and LM components
of \xmoverscore{}.

\subsection{\distilscore{}}\label{sec:distill}
\citet{Reimers2020MakingMS} show that the cosine between multilingual sentence
embeddings captures semantic similarity and can be used to assess cross-lingual
semantic textual similarity. Their approach to inducing embedding models is
based on multilingual knowledge distillation. We refer to this metric as
\distilscore. Their approach requires supervision at multiple levels. First,
parallel sentences are needed to induce multilingual models, and second, NLI
and STS corpora are required to induce teacher embeddings in the source
language.

\subsection{\sentsim{}}\label{sec:sentsim}
A key difference between \xmoverscore{} and \distilscore{} is that one approach
is based on word- and the other on sentence embeddings.
\citet{Song2021SentSimCS} and \citet{kaster-etal-2021-global} show that
combining approaches based on word-level and sentence-level representations can
substantially improve metrics. The metric of \citet{Song2021SentSimCS}, which
is called \sentsim, combines supervised \distilscore{} with one of two word
embedding-based metrics. The first one is quite similar to \xmoverscore{}, as
it is also based on WMD\@. The other one is a multilingual variant of
\bertscore{}~\citep{Zhang2020BERTScoreET}.

\section{Methods}
In this section, we introduce our fully unsupervised metric \metric{}.
\metric{} builds upon the existing metrics \xmoverscore, \distilscore, and
\sentsim{}, but eliminates all supervision signals and instead leverages the
dualities shown in Figure~\ref{fig:self}. In particular, we mine
pseudo-parallel data from unsupervised metrics, which we use (iteratively) to
(a) rectify deficient vector spaces (for \xmoverscore) and to (b) train
unsupervised MT systems which can generate pseudo-references (as
pseudo-references are in the same language as the hypothesis, this eliminates
problems of cross-lingual deficiency). Furthermore, we use pseudo-parallel data
to (c) induce an unsupervised sentence embedding model analogous to
\distilscore, which we can then (d) integrate with the unsupervised word based
model analogous to \sentsim{}. We now give details.

\subsection{\metric[wrd]}\label{sec:wmd}
\xmoverscore{} uses sentence-parallel data to extract word pairs for vector
space remapping. (i) We replace this parallel data with pseudo-parallel
data.\footnote{To extract the word pairs from the sentence-parallel data,
\xmoverscore{} uses \fastalign~\citep{Dyer2013ASF}, but since this depends
directly on how well sentences are aligned, we first replace it with
unsupervised \awesomealign~\citep{Dou2021WordAB} which only relies on
pre-trained language models.} (ii) In addition, we use pseudo-references to
address the issue of deficient vector spaces. We now give details on (i) and
(ii) below.

\begin{figure}[tb]
  \centering
  \scriptsize
  \input{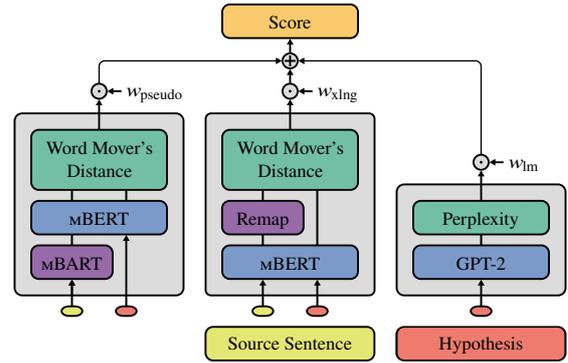}
  \caption{\metric[wrd] with pseudo
  references (left), unsupervised remapping (middle) and a LM
  (right).}%
  \label{fig:xmover-components}
\end{figure}

\paragraph{Efficient WMD Pseudo-Parallel Data  Mining}
Metrics such as \xmoverscore{} could in principle be used for pseudo-parallel
corpus mining since they can compare arbitrary sentences. However, when
WMD-based metrics are scaled to corpus mining, algorithmic efficiency problems
arise: (a) the computational complexity of WMD scales cubically with sentence
length~\citep{Kusner2015FromWE}; (b) to compare $m$ source to $m$ target
sentences, $m^2$ WMD invocations are necessary, which quickly becomes
intractable. Thus, we explore ways to improve the performance of WMD to mine
efficiently. In particular, \citet{Kusner2015FromWE} define a linear
approximation of WMD called \emph{word centroid distance}  (WCD) and a mining
algorithm that first sorts all target samples according to their WCD\@ to a
given query and computes exact WMD for the $k$ nearest neighbors. We use this
algorithm for efficient WMD-based pseudo-parallel data mining.

In our work, we apply this approach \emph{iteratively} (cf.\
Figure~\ref{fig:self}): we start out with an initial WMD metric (based on
\mbert{}), obtain \emph{sentence-level} pseudo-parallel data with it via the
efficient approximation algorithm described, and obtain a dictionary of word
pairs from unsupervised \awesomealign{} from the sentence pseudo-parallel data.
We use the pseudo-parallel word pairs with UMD and CLP to remap \mbert{}. From
this, we obtain a better WMD metric; then we iterate.

\paragraph{Pseudo References}
Apart from remapping, pseudo-parallel data could be used to overcome problems
of deficient vector spaces in other ways. Specifically, we want to mine enough
pseudo-parallel data to train an unsupervised MT system to translate source
sentences into the target language to create \emph{pseudo
references}~\citep{albrecht-hwa-2007-regression,gao-etal-2020-supert,Fomicheva2020MultiHypothesisMT}.
This would allow for a comparison with the hypothesis in the target language,
similar to reference-based metrics, circumventing alignment problems in
multilingual embeddings. This approach updates \metric[wrd] to
\begin{equation}
  \begin{split}\label{eq:wmd-combine}
    \m(x,y,y') ={}& \wxling\wmd^{(n)}(x,y) + \wlm\lm(y) \\ &+ \wpseudo\wmd(y,y'),\\
  \end{split}
\end{equation}
where $n$ denotes the iterations of remapping, and $\wpseudo$ is a new weight
to control the influence of the pseudo reference $y'$. All components of
\metric[wrd] are illustrated in Figure~\ref{fig:xmover-components}.

\subsection{\metric[snt]}\label{sec:cos}
Besides a word-based metric, we use pseudo-parallel data to induce an
unsupervised \emph{sentence level} metric, $\metric[snt]=\cos(x,y)$, based on
the cosine similarity between sentence embeddings. One could, similarly to
\distilscore{}, perform knowledge distillation but since our initial
experiments showed that this doesn't work well with pseudo-parallel data, we
chose another approach.

\paragraph{Contrastive Learning}
We explore contrastive learning for unsupervised \emph{multilingual} sentence
embedding induction, which has recently been successfully used to train
unsupervised \emph{monolingual} sentence embeddings~\citep{gao2021simcse}. In
our context, the basic idea  is to pull semantically close sentences together
and to push distant sentences apart in the embedding space. Let $x_i$ and $y_i$
be the embeddings of two sentences that are semantically related and $N$ an
arbitrary batch size. The training objective for this pair can be formulated as
\begin{equation}
  L_i=-\log\frac{\exp{(\frac{\cos(x_i, y_i)}{\tau}})}
    {\sum^N_{j=1,j\neq i}\exp({\frac{\cos(x_i, y_j)}{\tau}})},
\end{equation}
where $\tau$ is a temperature hyperparameter that can be used to either amplify
or dampen the assessed distances. For each sentence $x_i$, all remaining
sentences $y_{j\neq i}$ in the current batch  should be pushed apart in the
embedding space. For positive sentences that should be pulled together, we
again use pseudo-parallel sentence pairs. Since noisy data is beneficial for
contrastive learning~\citep{gao2021simcse}, we expect this paradigm to work
well with pseudo-parallel data. We use pooled \xlmr{} embeddings as sentence
representations, and, as with unsupervised remapping, we experiment with
multiple iterations of successive mining and sentence embedding induction
operations.

\paragraph{Ratio Margin Pseudo-Parallel Data  Mining}
As \metric[snt] is based on sentence embeddings, we cannot use the WMD-based
mining algorithm to obtain pseudo-parallel sentences since it requires access
to word-level representations. An alternative would be to just use cosine
similarity for mining, but that approach is susceptible to noise in the
data~\citep{Artetxe2019MarginbasedPC}. Instead, we follow
\citet{Artetxe2019MarginbasedPC} and use a ratio margin function defined as
\begin{equation}
  \margin(x, y)=\frac{%
    \cos(x,y)
  }{%
    \sum\limits_{z\in N_x}\frac{\cos(x,z)}{2k} +
    \sum\limits_{z\in N_y}\frac{\cos(y,z)}{2k}
  },
\end{equation}
where $N_x$ and $N_y$ are the $k$ nearest neighbors of sentence embeddings $x$
and $y$ in the respective language. Informally, this ratio margin function
divides the cosine similarities of the nearest neighbor by the average
similarities of the neighborhood.

\subsection{\metric[wrd\,$\oplus$\,snt]}\label{sec:ensembles}
Inspired by \sentsim{}, which combines word and sentence embeddings, we
similarly ensemble \metric[wrd] and \metric[snt]. We refer to this final metric
as \metric{} $\:\:=$ \metric[wrd\,$\oplus$\,snt] with two new weights $\wwmd$
and $\wcos=1-\wwmd$:
\begin{equation}
  \begin{split}
    \metric(x,y) ={}& \wwmd\operatorname{\metric[wrd]}(x,y)\\
    &+ \wcos\operatorname{\metric[snt]}(x,y).
  \end{split}
\end{equation}

\section{Experiments}
In this section, we evaluate all \metric{} variants at the segment
level\footnotemark{} and compare them to \type{1/2/3} upper bounds. We detail
additional hyperparameters in Appendix~\ref{sec:hyperparameters}.
\goodbreak\footnotetext{We do not evaluate at the system level since metrics
there often perform very similarly, making it difficult to determine the best
metric~\citep{mathur-etal-2020-results,freitag-etal-2021-results}.}

\subsection{Datasets}
We use various datasets to assess the performance of our metrics on
\begin{description*}[afterlabel={{, }}, itemjoin={{, and }}, after={{.}}]
  \item[MT evaluation] i.e., computing the correlation with human assessments
    using \pearson{} correlation
  \item[parallel sentence matching] a standard evaluation measure in the corpus
    mining field where a set of shuffled parallel sentences is searched to
    recover correct translation
    pairs~\citep{Guo2018EffectivePC,Kvapilkov2020UnsupervisedMS}. For this we
    report Precision at N (P@N)
\end{description*}

\paragraph{MT evaluation}
\begin{description*}[before={{In }}]
  \item[WMT-16 \textnormal{and} WMT-17,] each language pair consists of tuples
    of source sentences, hypotheses and references. Each tuple was annotated
    with a direct assessment (DA) score, which quantifies the adequacy of the
    hypothesis given the reference translation. Following \citet{Zhao2020OnTL}
    and \citet{Song2021SentSimCS}, we use these DA scores to assess the
    adequacy of the hypothesis given the source.
  \item[MLQE-PE] has been used in the WMT 2020 Shared Task on Quality
    Estimation~\citep{Specia2020FindingsOT}, and only provides source sentences
    and hypotheses for its language pairs, with no references. Each source
    sentence and hypothesis pair was annotated with cross-lingual direct
    assessment (CLDA) scores. In terms of annotation,
  \item[Eval4NLP] is very similar to MLQE-PE but focuses on non-English-centric
    language directions, especially de-zh and ru-de.
  \item[WMT-MQM] uses fine-grained error annotations from the Multidimensional
    Quality Metrics (MQM) framework~\citep{Freitag2021ExpertsEA} for adequacy
    assessments.  Like MLQE-PE and Eval4NLP, WMT-MQM also assigns scores based
    on source sentences and hypotheses.
\end{description*}
Additional statistics can be found in the appendix in
Table~\ref{tab:dataset-statistics}. Using ISO 639-1 codes, our datasets cover
the language pairs de-zh, ru-de, en-ru, en-zh, cs-en, de-en, en-de, et-en,
fi-en, lv-en, ne-en, ro-en, ru-en, si-en, tr-en, zh-en.

\paragraph{Parallel sentence matching}
To evaluate  on parallel sentence matching, we use the \textbf{News
Commentary}\footnote{\pref[http://]{data.statmt.org/news-commentary}} dataset. It
consists of parallel sentences crawled from economic and political data.

\subsection{Fine-grained analysis on de-en}\label{sec:preliminary}
To gain an understanding of the properties of the iterative techniques and the
influence of individual parameters / components, we conduct a fine-grained
analysis on the de-en language direction of WMT-16 (for MT evaluation) and News
Commentary v15 (for parallel sentence matching). We list examples of
pseudo-parallel data used during training in the appendix in
Table~\ref{table:pseudo-parallel}. The mined sentences are often semantically
similar, but contain factuality errors (e.g., have wrong places or numbers in
hypotheses).

\paragraph{Vector space remapping}
\label{par:remapping}
We explore if remapping works with  pseudo-parallel data. We use News Crawl for
mining. We randomly extract 40k monolingual sentences per language,  and select
the top 5\% sentence pairs with the highest metric scores for remapping. This
gives us the same number of sentences (2k pairs) as were used for remapping
\xmoverscore.

\begin{figure}[tb]
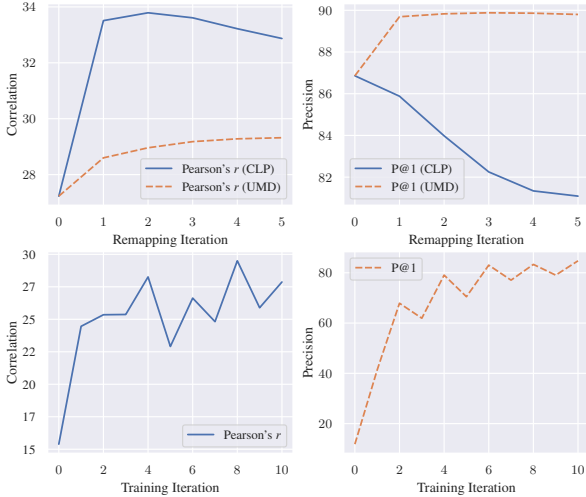

  \centering
  \input{remap-correlation.pgf}
  \input{remap-precision.pgf}
  \input{contrastive-correlation.pgf}
  \input{contrastive-precision.pgf}
  \caption{Results for unsupervised vector space remapping (top) and
  contrastive learning with \metric[snt] (bottom) for de-en. \pearson{} is
  computed on WMT-16 (MT evaluation) and P@1 on News Commentary v15 (parallel
  sentence matching).}%
  \label{fig:remap-contrastive}
\end{figure}

The results for UMD and CLP-based remapping on de-en can be seen in
Figure~\ref{fig:remap-contrastive} (top).
The figure contains two graphs, one for correlation with human judgments and
one for precision on parallel sentence matching. Each graph illustrates model
performance before remapping (Iteration 0) and after remapping one to five
times. After remapping once, both UMD and CLP improve substantially in
\pearson{} correlation. The improvement of CLP, however, is noticeably larger.
For subsequent iterations, UMD seems to continue to improve slightly, but the
correlations of CLP seem to drop. This can be explained by the results for
precision where the P@1 of CLP drops each iteration, meaning the remapping
capabilities of the metrics decrease. UMD does not exhibit this problem. Thus,
UMD could be a more robust choice for metrics that should perform reasonably
well on both tasks.

\begin{figure}[tb]
  \centering
  \input{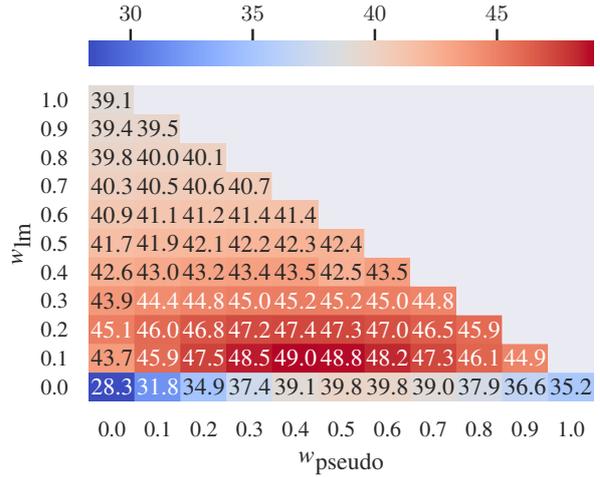}
  \caption{Influence of a language model and an MT system on \metric[wrd],
  segment-level \pearson{} for different values of $\wpseudo$ (weight for
  pseudo references) and $\wlm$ (weight for the language model) on WMT-16 de-en
  data. Note that the point $\wpseudo=\wlm=0$ uses only $\wmd(x,y)$; see
  Equation~\ref{eq:wmd-combine}. Here, $n=1$.}%
  \label{fig:lm}
\end{figure}

\paragraph{Pseudo References \& Language Model}\label{par:pseudo-references}
Next, we add a language model to the metric and investigate pseudo-parallel
corpus mining to train an MT system for pseudo references.
\citet{Tran2020CrosslingualRF} show that fine-tuning \mbart{} using
pseudo-parallel data leads to very promising results, so we use \mbart{} for
our own experiments as well. Since fine-tuning for MT is a very
resource-intensive undertaking requiring many parallel sentence
pairs~\citep{Barrault2020FindingsOT}, especially compared to our vector space
remapping experiments, we need considerably more training data. On average,
\citet{Tran2020CrosslingualRF} use around 200k pseudo-parallel sentence pairs
for training. To obtain the same amount with our extraction rate of 5\%, we now
use a pool of 4m sentences per language for mining. Our results on the de-en
data of WMT-16 are reported in Figure~\ref{fig:lm}, which is similar to an
ablation study.
On the x-axis, we vary the weight $\wpseudo \in \{0.0, 0.1,\dots,0.9,1.0\}$ for
\metric[wrd] with pseudo references, and on the y-axis, we explore different
weights $\wlm \in \{0.0, 0.1,\dots,0.9,1.0\}$ for the language model. We set
$\wxling=1-\wpseudo-\wlm$. The best correlation uses $\wpseudo=0.4$,
$\wlm=0.1$, and $\wxling=0.5$. The improvement when pseudo references and a
language model are included is substantial (over only using WMD)---e.g., we
improve from 28\% correlation with humans ($\wpseudo=\wlm=0$) to 49\% with the
best weight combination, an improvement of 75\%.

\paragraph{Contrastive Learning}
For \metric[snt], we also use 4m monolingual sentences per language for mining
but only retain the top 100k sentence pairs, as for the contrastive training
objective we additionally have to filter out duplicate sentences. The results
of \metric[snt] are shown in Figure~\ref{fig:remap-contrastive} (bottom). The
P@1 scores seem to steadily improve every two training iterations. Beginning
with the sixth iteration, the precision seems to converge.

\subsection{Other Languages: Results \& Analysis}
We now test our metrics on other languages and datasets. For \metric[snt], we
train its sentence embedding model for six iterations. For \metric[wrd], we
remap \mbert{} once with UMD and make use of a language model and pseudo
references obtained from an MT system. Based on Section~\ref{sec:preliminary},
we set $\wpseudo=0.4$, $\wlm=0.1$, and $\wxling=0.5$. Additionally, based on
analogous, unreported experiments, we set $\wwmd=0.6$ and $\wcos=0.4$ for
ensembling. Since determining the weights for \metric[wrd] and
\metric[wrd\,$\oplus$\,snt] this way constitutes a form of supervision, we also
evaluate weights chosen independently from our conducted experiments. Namely,
we also evaluate \metric[][+] with $\wwmd = \wcos = 0.5$. For $\wlm$, we follow
\xmoverscore{} and set it to $0.1$ (setting $\wlm$ lower makes sense because
the LM only addresses the hypothesis without considering the source);
accordingly, we set $\wxling = \wpseudo = 0.45$. Since $\wlm=0.1$ coincides
with our findings in Section~\ref{sec:preliminary}, we also evaluate
\metric[][++] where each component uses entirely uniform weights, i.e., $\wwmd
= \wcos = \frac{1}{2}$ and $\wlm = \wxling = \wpseudo = \frac{1}{3}$.

\begin{table*}[tb]
  \centering
  \begin{tabular}{clllllllll}
    \toprule
    & \textbf{Metric} & \textbf{WMT-16} & \textbf{WMT-17} & \textbf{MLQE-PE} & \textbf{Eval4NLP} & \textbf{WMT-MQM}\\
    \midrule\multirow{3.5}{*}{\makecell{\faint{\textbf{\type{1/2}}}\\\faint{\textbf{Supervised}}}}
    & \faint{\monotransquest} & \faint{64.68} & \faint{66.47} & \faint{66.28} & \faint{42.21} & \faint{42.23}\\
    & \faint{\cometqe}          & \faint{65.76} & \faint{68.77} & \faint{49.97} & \faint{40.00} & \faint{45.89}\\
    \arrayrulecolor{gray}\cmidrule(l){2-7}\arrayrulecolor{black}
    & \faint{\bleu}           & \faint{47.82} & \faint{47.01} & \faint{\---} & \faint{\---} &   \faint{19.76}\\
    \midrule\multirow{5}{*}{\makecell{\textbf{\type{3}}\\\textbf{Supervised}}}
    & \xmoverscore[UMD]   & 49.96  & 51.02 &            33.99 & 29.60 & 20.07\\
    & \xmoverscore[CLP]   & 53.10  &\underline{55.09}&            31.98 &\underline{36.93}& 20.59\\
    & \sentsim[BERTScore] & 51.86  &\underline{55.57}&    \textbf{45.36}& 26.60 & 13.71\\
    & \sentsim[WMD]       & 50.66  & 54.29 & \underline{44.72}& 24.44 & 12.24\\
    & \distilscore{}      & 43.79  & 51.22 & \underline{43.31}& 28.90 &  2.20\\
    \midrule\multirow{8}{*}{\textbf{Unsupervised}}
    & \metric[wrd]              &\underline{53.87}       & 53.52        & 31.13 & \underline{36.96}        &\textbf{23.28}\\
    & \metric[snt]                & 36.06 & 42.68 & 37.51 & 20.39 & -0.47\\
    & \metric[wrd\,$\oplus$\,snt] &\textbf{55.78}&\textbf{57.55}& 40.60 &\textbf{39.87}& 19.35\\
    \cmidrule(l){2-7}
    & \metric[wrd][+]             &\underline{53.66}& 52.97 & 30.94 &\underline{36.29} &\underline{23.15}\\
    & \metric[wrd\,$\oplus$\,snt][+] &\underline{55.28} &\underline{57.29}& 41.41 & \underline{38.22} & 17.71\\
    \cmidrule(l){2-7}
    & \metric[wrd][++]                & 51.88 & 51.73 & 23.76 & 25.42 & 19.00\\
    & \metric[wrd\,$\oplus$\,snt][++] &\underline{54.99}&\underline{56.49}& 34.15 & 32.05 & 17.08\\
    \bottomrule
  \end{tabular}
  \caption{Segment-level \pearson{} correlations with human judgments, averaged
  over language directions. The best results are highlighted in bold, while
  results that are not significantly worse (as determined by a two-sample
  t-test with 10\% significance level) are underlined.}%
  \label{tab:correlations}
\end{table*}
Correlations with human judgments averaged over language pairs are shown in
Table~\ref{tab:correlations} (individual results are in the appendix). We also
present the results of the popular \type{2} metric \bleu, where possible, and
the recent \type{1} metrics
\monotransquest~\citep{transquest2020a,transquest2020b} and
\cometqe~\citep{rei-etal-2021-references}. Finally, as more direct competitors,
we compare to the \type{3} metrics \xmoverscore{}, \sentsim{}, and
\distilscore. We compute all reported scores ourselves.

Overall, the tuned weights of \metric[] perform marginally better than
\metric[][+] on most datasets, but \metric[][+] is usually a very close second.
\metric[][++] performs worse, however, and only competitively on two of the
five datasets. This indicates that the language model should be set to a lower
value, a choice that makes intuitively sense.

Expectedly, \distilscore{}, which uses parallel data, is always better than
\metric[snt], which uses pseudo-parallel data. In contrast, \metric[wrd] is
generally on par with \xmoverscore{}, even though \xmoverscore{} uses real
parallel data---the difference is that \metric[wrd] also leverages
pseudo-references which \xmoverscore{} does not. Indeed, from
Figure~\ref{fig:lm}, we observe that the pseudo-references can make an
improvement of up to 1--11 points in correlation (comparing `column' labeled
$w_{\text{pseudo}}=0$ to the columns $w_{\text{pseudo}}>0$).

Our metrics beat reference-based \type{2} \bleu{} across the board. \type{1}
metrics, which are fine-tuned on human scores, are generally the best.
Intriguingly, the only two language pairs where our metrics are on par with
them are the non-English de-zh and ru-de from Eval4NLP\@. These languages are
outside the training scope of the current \type{1} metrics and thus test their
generalization abilities. For example, on ru-de our best metric outperforms
\monotransquest{} by 5 points correlation and \cometqe{} by 9 points
(Table~\ref{tab:mqm-newseval4nlp-pearson} in the appendix).

\metric[] and \metric[][+] also outperform the \type{3} upper bounds on four of
five datasets. On WMT-16, WMT-17 and Eval4NLP, they have the best overall
results. On WMT-MQM, \metric[wrd] alone is best. The drop in performance for
the combined metric is caused by \metric[snt], which on its own performs very
badly. As supervised \distilscore{} exhibits the same issues, this could be a
general problem for sentence embeddings based metrics on this dataset. We
identify further reasons in Appendix~\ref{sec:supplementary}.

For MLQE-PE, the \sentsim{} metrics perform best on average among \type{3} and
our metrics---although our reproduced scores for this dataset differ noticeably
from the authors' results, due to issues in their original
code~\citep{Chen2022ReproducibilityIF}. Among our self-learned metrics, the
combined variant performs best on average again, but still is 3--5 points below
\sentsim{} and \distilscore{}, even though it outperforms both \xmoverscore{}
variants by over 6 points. Interestingly, \metric[snt] works better than
\metric[wrd], unlike for the other datasets. Similarly, \distilscore{} clearly
outperforms \xmoverscore{}. This could be because MLQE-PE contains Sinhala, a
language \mbert{} was not trained on. Another explanation is the data
collection scheme for ru-en, which uses different sources of parallel
sentences, mainly colloquial data and Russian proverbs, which use rather
unconventional grammar~\citep{Fomicheva2020MLQEPEAM}. This apparently confuses
the language model and MT system which have been trained on data from other
domains. When we exclude si-en and ru-en from MLQE-PE,
\metric[wrd\,$\oplus$\,snt] performs best, with an average \pearson{} of 44.22
for tuned weights and 44.45 for default weights vs.\ 43.82 for
\sentsim[BERTScore]. In Appendix~\ref{sec:finetune}, we show that incorporating
real parallel data (in addition to pseudo-parallel data) at an order of
magnitude lower than \sentsim{} allows us to outperform \sentsim{} on MLQE-PE
also.

\section{Discussion}\label{sec:analysis}
Throughout, we have presented a mix of results and analysis, which we now
summarize and discuss. In Figure~\ref{fig:lm}, we conducted an ablation study
on the individual components of \metric[wrd] (on de-en). This showed that all
three components (pseudo-references, language model, WMD) matter; by itself,
the LM is more important than the pseudo-references which are more important
than WMD\@. However, in combination, the LM is least important. We also showed
that pseudo-parallel data can successfully rectify deficient multilingual
vector spaces, similar to real parallel data, see
Figure~\ref{fig:remap-contrastive}. We note, however, that pseudo-parallel data
may introduce an important bias in our data sampling: namely, it may mine
factually incorrect parallel sentences, see Table~\ref{table:pseudo-parallel}
in the appendix, which may amplify issues of adversarial robustness; see our
discussion in Section~\ref{sec:limitations}.

We remark that, depending on the annotation scheme, better correlations with
human judgments do not necessarily entail better
metrics~\citep{Freitag2021ExpertsEA}. The datasets in this work were annotated
using either DA, CLDA, or MQM scores, with MQM explicitly addressing this
problem. Since our metrics are consistent regardless of the annotation scheme,
they are unlikely to overfit a particular one.

We finally note that combining word and sentence-level models is meaningful,
because they offer complementary views \citep{Song2021SentSimCS}.
\citet{kaster-etal-2021-global} also show that they capture orthogonal
linguistic factors to varying degrees. Such complementarity may also stem from
different underlying vector spaces, i.e., \mbert{} vs.\ \xlmr{} that we use in
sentence- and word-level metrics.

\section{Related Work}
All metrics in this work presented so far treated the MT model generating the
hypotheses as a black-box. There also exists a recent line of work of so-called
glass-box metrics, which actively incorporate the MT model under test into the
scoring
process~\citep{Fomicheva2020UnsupervisedQE,Fomicheva2020MultiHypothesisMT}. In
particular, \citet{Fomicheva2020MultiHypothesisMT} explore whether the MT model
under test can be used to generate additional hypotheses
\citep{Dreyer2012HyTERMS}. A crucial difference to our metrics is the required
availability of the original MT model, which we are agnostic about. The MT
models used in \citet{Fomicheva2020MultiHypothesisMT} are all trained on
parallel data, which makes their approach a supervised metric in our sense.

Other recent metrics that leverage the relationship between metrics and (MT)
systems are \prism~\citep{Thompson2020AutomaticMT} and
\bartscore~\citep{yuan2021bartscore}. We do not classify them as unsupervised,
however, as \prism{} is trained from scratch on parallel data and \bartscore{}
uses a \bart{} model fine-tuned on labeled summarization or paraphrasing
datasets.

There are also multilingual sentence embedding models which are highly relevant
in our context. \citet{Kvapilkov2020UnsupervisedMS}, for example, fine-tune
\xlmr{} on synthetic data translated with an unsupervised MT system. Similar to
our contrastive learning approach, the resulting embedding model is completely
unsupervised. Important differences are that our sentence embedding model can
be improved iteratively and does not rely on an MT system. We leave a
comparison to future work.

Finally, the idea of fully unsupervised text generation systems has originated
in the MT
community~\citep{artetxe2018unsupervised,lample-etal-2018-phrase,artetxe-etal-2019-effective}.
Given the similarity of MT systems and evaluation metrics, designing fully
unsupervised evaluation metrics is an apparent next step, which we take in this
work.

\section{Conclusion}\label{sec:conclusion}
In this work, we aimed for sample efficient evaluation metrics that do not use
any form of supervision. In addition, our novel metrics should be maximally
effective, i.e., of high quality. To achieve this, we leveraged pseudo-parallel
data obtained from fully unsupervised evaluation metrics in three ways: we (i)
remapped deficient vector spaces using the pseudo-parallel data, (ii) trained
an unsupervised MT system from it (yielding pseudo references), and (iii)
induced unsupervised multilingual sentence embeddings. To enable our approach,
we also explored \emph{efficient} pseudo-parallel corpus mining algorithms
based on our metrics as an orthogonal contribution. Finally, we showed that our
approach is effective and can outperform three supervised upper bounds (making
use of parallel data) on 4 out of 5 datasets we included in our comparison.

In future work, we want to aim for algorithmic efficiency, include \emph{pseudo
source texts} as additional components (using the MT system in backward
translation), and address the missing dualities discussed in
Figure~\ref{fig:self} (i.e., use of metrics as optimization criteria and MT
systems to generate additional pseudo-parallel data). Further, our approach has
substantial room for improvement given that we selected hyperparameters
completely unsupervised or based on one high-resource language pair (de-en).
Thus, it will be particularly intriguing to explore  \emph{weakly-supervised
approaches} which leverage minimal forms of supervision.

\section{Limitations}\label{sec:limitations}
Limitations of our metrics include (1) \emph{algorithmic inefficiency}, (2)
\emph{resource inefficiency}, (3) the brittleness of unsupervised MT systems in
certain situations, and (4) issues of adversarial robustness.

(1) Some of the components of \metric[wrd] (mainly the MT system) have high
computational costs. For example, \xmoverscore{} and \sentsim[BERTScore] take
less than 30 seconds to score 1000 hypotheses on an Nvidia V100 GPU\@.
\metric[wrd], on the other hand, takes over 2.5 minutes. This algorithmic
inefficiency trades off with our sample efficiency, by which we did not use any
supervision signals. In future work, we aim to experiment with efficient MT
architectures  to reduce computational costs
\citep{kamal-eddine-etal-2022-frugalscore,Grunwald2022CanWD}.

(2) Similarly to \xmoverscore{}, \monotransquest{} or \sentsim[], our metrics
use high-quality encoders such as \mbert{}, which are not only memory and
inference inefficient but also leverage large monolingual resources. Future
work should thus not only investigate using smaller \mbert{} models but also
models that leverage smaller amounts of monolingual resources.
\citet{wang-etal-2020-extending}, for example, propose a competitive LSTM-based
approach that completely forgoes monolingual resources and instead uses small
parallel corpora (i.e., a few hundred parallel sentences as a weak supervision
signal). Similarly, we give a recipe for improving \mbert{} for unseen
languages using limited amounts of parallel data in
Appendix~\ref{sec:finetune}.

(3)  Using unsupervised MT approaches, as we do via pseudo references, may be
less effective for truly low-resource languages
\citep{marchisio-etal-2020-unsupervised}. However, this remains a very active
research field with a constant influx of more powerful solutions
\citep{ranathunga2021neural,sun2021unsupervised}.

(4) As indicated in Sections~\ref{sec:preliminary}~and~\ref{sec:analysis}, our
mined pseudo-parallel data tends to contain factual inconsistencies such as
``Uruguay was seventh'' vs.\ (a translation of) ``Russia was second''. As a
consequence, our induced metrics may be less robust than existing metrics
\citep{Chen2022MENLIRE,rony-etal-2022-rome}. An approach to address this
inconsistency would be to retain only high probability aligned words in
parallel sentences (recall that we infer word-level parallel data from
sentence-level parallel data).

\section{Acknowledgments}
We thank all reviewers for their valuable feedback, hard work, and time. The
last author was supported by DFG grant EG 375/5--1.

\bibliographystyle{acl_natbib}
\bibliography{literature}

\clearpage
\appendix
\section{Vector Space Remapping}\label{sec:remapping}
\citet{Zhao2020OnTL} explore two different remapping approaches for
\xmoverscore, which are defined as follows:
\begin{description}
  \item[Procrustes alignment] \citet{Mikolov2013ExploitingSA} propose to
    compute a linear transformation matrix $\mathbf{W}$ which can be used to
    map a vector $x$ of a source word into the target language subspace by
    computing $\mathbf{W}x$. The transformation can be computed by solving the
    problem \begin{equation} \min_\mathbf{W}||\mathbf{WX} - \mathbf{Y}||_2.
    \end{equation} Here $\mathbf{X}, \mathbf{Y}$ are matrices with embeddings
    of source and target words, respectively, where the tuples $(x_i,y_i)\in
    \mathbf{X}\times \mathbf{Y}$ come from parallel word pairs. \xmoverscore{}
    constrains $\mathbf{W}$ to be an orthogonal matrix such that
    $\mathbf{W}^{\intercal}\mathbf{W}=\mathbf{I}$, since this can lead to
    further improvements~\citep{Xing2015NormalizedWE}. \citet{Zhao2020OnTL}
    call this remapping Linear Cross-Lingual Projection remapping (CLP).

  \item[De-biasing] The second remapping method of \xmoverscore{} is
    rooted in the removal of biases from word embeddings.
    \citet{Dev2019AttenuatingBI} explore a bias attenuation technique called
    Universal Language Mismatch-Direction (UMD). It involves a bias vector
    $v_B$, which is supposed to capture the bias direction. For each word
    embedding $e$, an updated word embedding $e'$ is computing by subtracting
    their projections onto $v_B$, as in \begin{equation} e' = e - (e\cdot
    v_B)v_B, \end{equation} where $\cdot$ is the dot product. To obtain the
    bias vector $v_B$, \citet{Dev2019AttenuatingBI} use a set $\mathcal{E}$ of
    word pairs that should be de-biased (e.,g.\ \emph{man} and \emph{woman}).
    The subtractions of the embeddings of the words in each pair are then
    stacked to form a matrix $\mathbf{Q}$, and the bias vector $v_B$ is its
    top-left singular vector. \citet{Zhao2020OnTL} use the same approach for
    \xmoverscore{}, but $\mathcal{E}$ instead consists of parallel word pairs.
\end{description}

\citet{Zhao2020OnTL} show that these remapping methods lead to substantial
improvements of their \xmoverscore{} metric (on average, up to 10 points in
correlation). The required parallel word pairs were extracted from sentences of
the EuroParl corpus~\citep{Koehn2005EuroparlAP} using the
\fastalign{}~\citep{Dyer2013ASF} word alignment tool. The best results were
obtained when remapping on 2k parallel sentences.

\section{Filtering}\label{sec:filtering}
Since large corpora tend to include low-quality data points, we follow
\citet{Artetxe2019MarginbasedPC} and \citet{Keung2021UnsupervisedBM} and apply
three simple filtering techniques. We first remove all sentences from each
monolingual corpus for which the \fasttext{} language identification
tool~\citep{joulin2016bag} predicts a different language. We then filter all
sentences which are shorter than 3 tokens or longer than 30 tokens. As the last
step, we discard sentence pairs sharing substantial lexical overlap, which
prevents degenerate alignments of, e.g., proper names. We remove all sentence
pairs for which the Levenshtein distance detects an overlap of over 50\%.

\section{Fine-Tuning on Parallel Data}\label{sec:finetune} 
\begin{figure}[tb]
  \centering
    \input{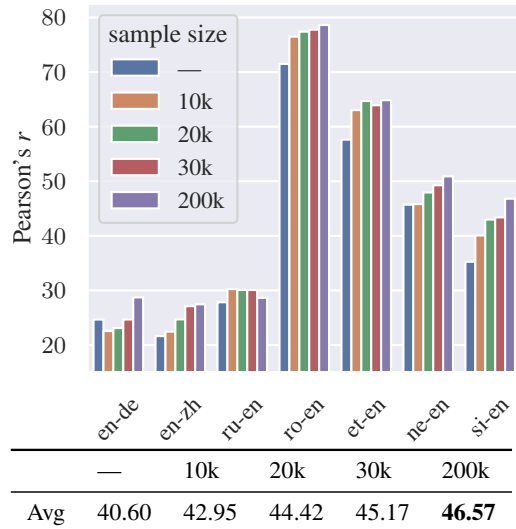}
    {\footnotesize
      \begin{tabular}{llllll}
        \toprule
            & ---   & 10k   & 20k   & 30k   & 200k\\
        \midrule
        Avg & 40.60 & 42.95 & 44.42 & 45.17 & \textbf{46.57}\\
        \bottomrule
      \end{tabular}
    }
    \caption{\pearson{} correlations on MLQE-PE for \metric[wrd\,$\oplus$\,snt]
    when fine-tuning on limited amounts of parallel data. We explore sample
    sizes of 10k, 20k, 30k, and 200k.}%
    \label{fig:finetune}
\end{figure}
To examine whether and by how much we can further improve our metrics using
forms of supervision, we experiment with a fine-tuning step on \emph{parallel
sentences} and treat self-learning on pseudo-parallel data as
pre-training~\citep{He2020RevisitingSF}. We use the parallel data to fine-tune
the contrastive sentence embeddings of \metric[snt] and the MT system of
\metric[wrd], which is responsible for generating pseudo references. Further,
we also compute new remapping matrices for \metric[wrd]. Since CLP is superior
to UMD when parallel data is used (see Section~\ref{par:remapping}), we compute
these remapping matrices using CLP instead of UMD\@. To assess how different
amounts of parallel sentences affect performance, we fine-tune our metrics on
10k, 20k, 30k, and 200k parallel sentences. We use
WikiMatrix~\citep{Schwenk2021WikiMatrixM1} and the Nepali Translation Parallel
Corpus~\citep{Duwal2019EffortsIT} to obtain parallel sentences.

\pearson{} correlations with human judgments for individual and averaged
language pairs are shown in Figure~\ref{fig:finetune}; we focus on MLPQE-PE,
where our metrics performed worst. Overall, introducing parallel data into the
training process consistently improves performance for the majority of language
directions; more parallel data leads to better results. The relatively biggest
improvements are achieved for the si-en language direction, which is in
accordance with our discussion above. When fine-tuning with 30k parallel
sentences, the performance of our metrics is roughly on par with the \sentsim{}
variants (see Table~\ref{tab:correlations}). With 200k parallel sentences, our
metrics clearly outperform \sentsim{}, which uses millions of parallel
sentences and NLI data as supervision signals.

\section{Hyperparameters}\label{sec:hyperparameters}
For efficient WMD pseudo-parallel mining, we set $k=20$ for remapping, and
$k=1$ for training \mbart. For ratio-margin-based pseudo-parallel mining, we
use $k=5$.
With regard to training \metric[snt], we follow \citet{gao2021simcse}, and
iteratively train \xlmr{} for one training epoch with a learning rate of 5e-5,
a batch size of 256, and a temperature coefficient of $\tau=0.05$ utilizing the
AdamW optimizer~\citep{Loshchilov2019adamw}.
Fine-tuning of \mbart{} was performed for three epochs with a batch size of
four and using the same learning rate of 5e-5 as well as AdamW optimizer.

We decided to continue using \mbert{} in \metric[wrd] for two reasons. Firstly,
we want \metric[wrd] to remain directly comparable to XMoverScore which is
based on \mbert. Secondly, in our own experience, vanilla \mbert{} is very
robust in terms of layer choice, especially compared to vanilla \xlmr.
Intuitive choices like the first or last layer work very well for a lot of
problems. This is an important property for unsupervised metrics, since we
can’t easily justify a (supervised) hyperparameter search in an unsupervised
setting to determine the best layer or even a linear combination of those.

\section{Supplementary Data and Results}\label{sec:supplementary}
Table~\ref{table:pseudo-parallel} shows examples of pseudo-parallel data
obtained with \metric[wrd] and \metric[snt].
Tables~\ref{tab:newstest-2016-pearson},~\ref{tab:newstest-2017-pearson},~%
\ref{tab:mlqe-pearson}, and~\ref{tab:mqm-newseval4nlp-pearson} show
segment-level \pearson{} correlations with human judgments on WMT-16, WMT-17,
MLQE-PE, as well as WMT-MQM and Eval4NLP, respectively.
Table~\ref{tab:dataset-statistics} provides additional statistics for each
dataset.

A surprising finding is the poor performance of \metric[snt] and \distilscore{}
on the German-English language pair in
Table~\ref{tab:mqm-newseval4nlp-pearson}. It is well known that high-resource
language directions, such as English-German, can be affected by a lack of
low-quality translations~\citep{Fomicheva2020MLQEPEAM}, and with only
high-quality translations available, there is little variation in the scores,
which makes a meaningful assessment of correlation
difficult~\citep{Specia2020FindingsOT}. Further, since both \metric[snt] and
\distilscore{} are based on sentence embeddings of \xlmr{}, and sentence
embedding-based metrics are known to be a bit worse on average than their word
embedding-based counterparts, we believe that both aspects combined could be
the root cause for this.

\begin{table*}[htb]
    \centering
    \begin{tabularx}{\textwidth}{lXX}
    \toprule
         \textbf{Case} & \textbf{Source} & \textbf{Target}\\ \midrule
         Top-WRD & Uruguay belegt mit vier Punkten nur Platz Sieben. & Russia was second with four gold and 13 medals. \\
         Top-WRD & Soweit lautet zumindest die Theorie. & That, at least, is the theory.\\
         Rnd-WRD & Die USA stellen etwa 17.000 der insgesamt 47.000 ausländischen Soldaten in Afghanistan. & Currently, there are about 170,000 U.S. troops in Iraq and 26,000 in Afghanistan.\\
         Rnd-WRD & ``Das ist eine schwierige Situation'', sagte Kaczynski. & ``It seemed like a ridiculous situation,'' Vanderjagt said.\\
         \midrule
         Top-SNT & Die Wahlen für ein neues Parlament sollen dann Anfang Januar stattfinden. & 	Parliamentary elections are to be held by January.\\
         Top-SNT & Anzeichen für die Blauzungenkrankheit sind Fieber, Entzündungen und Blutungen an der Zunge der Tiere.	& Contact with the creatures can cause itching, rashes, conjunctivitis and, in some cases, breathing problems.\\
         Rnd-SNT & Riesen-Wirbel an der Universität Zagreb: An der wirtschaftlichen Fakultät und am Institut für Verkehrsstudien durchsuchen Polizisten die Büros von Dozenten.	& Those attending the Soil Forensics International Conference work in the fields of science, policing, forensic services as well as private industries.\\
         Rnd-SNT & Frankfurt soll WM-Finale der Frauen ausrichten &	The women's tournament gets underway on Sunday.\\
         \bottomrule
    \end{tabularx}
    \caption{Pseudo-parallel data obtained via \metric[wrd] and \metric[snt];
      top and random sentence pairs. The mined sentences are semantically
      similar, but contain factuality errors (e.g., have wrong places or
      numbers in hypotheses).}%
    \label{table:pseudo-parallel}
\end{table*}

\begin{table*}[htb]
  \centering
  \begin{tabular}{cllllllll}
    \toprule
    & \textbf{Metric} & de-en & en-ru & ru-en & ro-en & cs-en & fi-en & tr-en\\
    \midrule\multirow{3.5}{*}{\makecell{\faint{\textbf{\type{1/2}}}\\\faint{\textbf{Supervised}}}}
    & \faint{\monotransquest} & \faint{61.65} & \faint{66.69} & \faint{63.32} & \faint{62.36} & \faint{67.67} & \faint{68.33} & \faint{62.71}\\
    & \faint{\cometqe}          & \faint{65.73} & \faint{71.91} & \faint{69.71} & \faint{66.40} & \faint{67.98} & \faint{64.83} & \faint{53.73}\\
    \arrayrulecolor{gray}\cmidrule(l){2-9}\arrayrulecolor{black}
    & \faint{\bleu}           & \faint{45.39} & \faint{55.08} & \faint{46.33} & \faint{47.09} & \faint{53.80} & \faint{39.92} & \faint{47.15}\\

    \midrule\multirow{5}{*}{\makecell{\textbf{\type{3}}\\\textbf{Supervised}}}
    & \xmoverscore[UMD]   & 43.46 & 62.16 &\textbf{60.52}& 47.88 & 58.83 & 43.52 & 33.34\\
    & \xmoverscore[CLP]   & 45.29 & 63.58 & 56.12 & 54.24 & 58.89 & 51.40 & 42.14\\
    & \sentsim[BERTScore] & 48.49 & 50.37 & 57.89 & 55.06 & 59.08 & 46.66 &\textbf{45.47}\\
    & \sentsim[WMD]       & 47.78 & 48.49 & 56.19 & 54.48 & 56.87 & 46.00 &44.84\\
    & \distilscore{}      & 40.62 & 41.21 & 43.16 & 51.22 & 49.51 & 39.65 & 41.14\\

    \midrule\multirow{8}{*}{\textbf{Unsupervised}}
    & \metric[wrd]                & 48.94 & 60.97 & 59.09 & 56.06 & 57.15 & 53.76 & 41.15\\
    & \metric[snt]                & 30.48 & 39.73 & 35.31 & 44.18 & 40.80 & 31.30 & 30.62\\
    & \metric[wrd\,$\oplus$\,snt] &\textbf{50.37}& 62.92 & 60.49 & 60.07 &\textbf{59.25}& 52.93 & 44.41\\
    \cmidrule(l){2-9}
    & \metric[wrd][+]                & 48.97 & 60.57 & 58.35 & 56.23 & 56.44 & 53.78 & 41.30\\
    & \metric[wrd\,$\oplus$\,snt][+] & 50.29 & 62.37 & 59.43 &\textbf{60.71}& 58.47 & 51.48 & 44.19\\
    \cmidrule(l){2-9}
    & \metric[wrd][++]                & 44.50 & 61.45 & 53.50 & 56.03 & 52.95 & 54.08 & 40.65\\
    & \metric[wrd\,$\oplus$\,snt][++] & 47.27 &\textbf{64.14}& 56.40 & 60.36 & 56.29 &\textbf{55.62}& 44.85 \\
    \bottomrule
  \end{tabular}
  \caption{Segment-level \pearson{} correlations with human judgments on the WMT-16 dataset.}%
  \label{tab:newstest-2016-pearson}
\end{table*}

\begin{table*}[htb]
  \centering
  \begin{tabular}{cllllllll}
    \toprule
    & \textbf{Metric} & cs-en & de-en & fi-en & lv-en & ru-en & tr-en & zh-en\\
    \midrule\multirow{3.5}{*}{\makecell{\faint{\textbf{\type{1/2}}}\\\faint{\textbf{Supervised}}}}
    & \faint{\monotransquest} & \faint{60.93} & \faint{63.54} & \faint{65.33} & \faint{72.01} & \faint{59.56} & \faint{75.91} & \faint{68.03}\\
    & \faint{\cometqe}          & \faint{68.99} & \faint{69.34} & \faint{72.83} & \faint{64.75} & \faint{69.06} & \faint{68.43} & \faint{68.01}\\
    \arrayrulecolor{gray}\cmidrule(l){2-9}\arrayrulecolor{black}
    & \faint{\bleu}           & \faint{41.22} & \faint{41.29} & \faint{56.48} & \faint{39.28} & \faint{45.99} & \faint{53.06} & \faint{51.75}\\

    \midrule\multirow{5}{*}{\makecell{\textbf{\type{3}}\\\textbf{Supervised}}}
    & \xmoverscore[UMD]   & 41.72 & 51.19 &        56.61 & 56.11 &        47.24 & 46.52 & 57.73\\
    & \xmoverscore[CLP]   & 47.76 & 50.04 & 62.22 & 63.95 &        48.79 & 52.97 & 59.88 \\
    & \sentsim[BERTScore] & 49.90 & 52.26 &        57.85 & 57.42 &\textbf{55.10}& 56.84 & 59.59\\
    & \sentsim[WMD]       & 47.62 & 50.42 &        56.59 & 56.91 &        53.42 & 56.24 & 58.86\\
    & \distilscore{}      & 46.42 & 45.64 &        54.03 & 55.51 &        54.13 & 54.04 & 50.89\\

    \midrule\multirow{8}{*}{\textbf{Unsupervised}}
    & \metric[wrd]                & 46.70 & 52.71 & 61.91 & 59.22 & 49.10 & 50.06 & 54.95\\
    & \metric[snt]                & 38.89 & 41.92 & 39.77 & 48.95 & 37.27 & 48.83 & 43.10\\
    & \metric[wrd\,$\oplus$\,snt] &\textbf{50.67}&\textbf{56.50}& 61.86 & 64.54 & 53.17 & 57.31 & 58.80\\
    \cmidrule(l){2-9}
    & \metric[wrd][+]                & 44.55 & 52.96 & 60.96 & 59.02 & 48.82 & 49.70 & 54.75\\
    & \metric[wrd\,$\oplus$\,snt][+] & 48.62 & 56.31 & 60.44 & 63.62 & 54.19 &\textbf{57.54}&\textbf{60.28}\\
    \cmidrule(l){2-9}
    & \metric[wrd][++]                & 45.41 & 48.00 & 61.33 & 60.64 & 49.65 & 47.06 & 50.00\\
    & \metric[wrd\,$\oplus$\,snt][++] & 49.27 & 52.20 &\textbf{63.10}&\textbf{66.64}& 53.70 & 54.93 & 55.62\\
    \bottomrule
  \end{tabular}
  \caption{Segment-level \pearson{} correlations with human judgments on the WMT-17 dataset.}%
  \label{tab:newstest-2017-pearson}
\end{table*}

\begin{table*}[htb]
  \centering
  \begin{tabular}{cllllllll}
    \toprule
    & \textbf{Metric} & en-de & en-zh & ru-en & ro-en & et-en & ne-en & si-en\\
    \midrule\multirow{2}{*}{\makecell{\faint{\textbf{\type{1}}}\\\faint{\textbf{Supervised}}}}
    & \faint{\monotransquest} & \faint{41.85} & \faint{45.76} & \faint{76.76} & \faint{88.81} & \faint{73.19} & \faint{75.70} & \faint{61.90}\\
    & \faint{\cometqe}          & \faint{36.03} & \faint{30.70} & \faint{49.33} & \faint{64.95} & \faint{63.45} & \faint{57.46} & \faint{47.85}\\

    \midrule\multirow{5}{*}{\makecell{\textbf{\type{3}}\\\textbf{Supervised}}}
    & \xmoverscore[UMD]   &        16.56 &        16.48  &        28.07  &        65.83 &        53.95 &        38.23 &        18.81 \\
    & \xmoverscore[CLP]   & 25.59 &        20.25  &        20.31  &        57.34 &\textbf{58.46}&        25.15 &        16.74 \\
    & \sentsim[BERTScore] &         6.15 &        22.23  & 47.30 &\textbf{78.55}&        55.09 &\textbf{57.09}&\textbf{51.14}\\
    & \sentsim[WMD]       &         3.86 & 22.62 &\textbf{47.46} & 77.72 &        54.60 & 57.00 & 49.79 \\
    & \distilscore{}      &        12.96 &\textbf{28.68} &        45.34  &        76.57 &        51.16 &        46.73 &        41.76 \\

    \midrule\multirow{8}{*}{\textbf{Unsupervised}}
    & \metric[wrd]                & 24.53 & 21.58 & 16.66 & 60.30 & 54.83 & 28.62 & 11.38\\
    & \metric[snt]                & 13.62 & 13.27 & 39.09 & 66.50 & 41.75 & 46.83 & 41.49\\
    & \metric[wrd\,$\oplus$\,snt] & 24.67 & 21.63 & 27.83 & 71.48 & 57.62 & 45.70 & 35.25\\
    \cmidrule(l){2-9}
    & \metric[wrd][+]                & 23.78 & 21.40 & 15.18 & 60.39 & 56.27 & 27.34 & 12.22\\
    & \metric[wrd\,$\oplus$\,snt][+] & 22.00 & 21.41 & 27.71 & 73.65 & 57.94 & 47.27 & 39.91\\
    \cmidrule(l){2-9}
    & \metric[wrd][++]                & 26.17 & 27.16 & 0.75 & 32.92 & 52.24 & 12.79 & 14.30\\
    & \metric[wrd\,$\oplus$\,snt][++] &\textbf{26.37}& 27.26 & 9.29 & 57.70 & 56.81 & 31.24 & 30.38\\
    \bottomrule
  \end{tabular}
  \caption{Segment-level \pearson{} correlations with human judgments on the MLQE-PE dataset.}%
  \label{tab:mlqe-pearson}
\end{table*}

\begin{table*}[htb]
  \centering
  \begin{tabular}{clllll}
    \toprule
    & & \multicolumn{2}{c}{\textbf{WMT-MQM}} & \multicolumn{2}{c}{\textbf{Eval4NLP}}\\
    \cmidrule(lr){3-4}\cmidrule(lr){5-6}
    & \textbf{Metric} & en-de & zh-en & de-zh & ru-de\\
    \midrule\multirow{3.5}{*}{\makecell{\faint{\textbf{\type{1/2}}}\\\faint{\textbf{Supervised}}}}
    & \faint{\monotransquest} & \faint{31.38} & \faint{53.07} & \faint{34.66} & \faint{49.76}\\
    & \faint{\cometqe}          & \faint{40.94} & \faint{50.84} & \faint{33.51} & \faint{46.49}\\
    \arrayrulecolor{gray}\cmidrule(l){2-6}\arrayrulecolor{black}
    & \faint{\bleu}           & \faint{17.94} & \faint{21.58} & \faint{ --- } & \faint{ --- }\\

    \midrule\multirow{5}{*}{\makecell{\textbf{\type{3}}\\\textbf{Supervised}}}
    & \xmoverscore[UMD]   & 12.72 & 27.41 &  10.85 &        48.35 \\
    & \xmoverscore[CLP]   & 12.89 & 28.29 &  21.61 &        52.25 \\
    & \sentsim[BERTScore] &  1.03 & 26.39 &  -7.71 &\textbf{60.91}\\
    & \sentsim[WMD]       & -0.23 & 24.70 & -11.54 & 60.42 \\
    & \distilscore{}      & -2.66 &  7.06 &   6.57 &        51.22 \\

    \midrule\multirow{8}{*}{\textbf{Unsupervised}}
    & \metric[wrd]                & 18.13 &\textbf{28.43}& 29.29 & 44.63\\
    & \metric[snt]                & -5.93 &  4.99 &  0.86 & 39.91\\
    & \metric[wrd\,$\oplus$\,snt] &        13.94 &        25.17 &        24.66 & 55.08\\
    \cmidrule(l){2-6}
    & \metric[wrd][+]                & 19.83 & 26.46 & 29.99 & 42.59\\
    & \metric[wrd\,$\oplus$\,snt][+] & 13.72 & 21.7  & 22.64 & 53.8 \\
    \cmidrule(l){2-6}
    & \metric[wrd][++]                &\textbf{20.19}& 17.80 &\textbf{36.51}& 14.32\\
    & \metric[wrd\,$\oplus$\,snt][++] & 17.18 & 16.97 & 34.00 & 30.09\\
    \bottomrule
  \end{tabular}
  \caption{Segment-level \pearson{} correlations with human judgments on the WMT-MQM and Eval4NLP datasets.}%
  \label{tab:mqm-newseval4nlp-pearson}
\end{table*}

\begin{table*}[htb]
  \centering
  \begin{tabular}{lllll}
  \begin{tabular}{lllll}
    \toprule
    Dataset  & Pairs & Tokens & Type\\
    \midrule
      WMT-16 &   560 &   9886 &   CA\\
      WMT-17 &   560 &  11189 &   CA\\
     MLQE-PE &  1000 &  15263 & CLDA\\
    Eval4NLP &  1295 &  25975 & CLDA\\
     WMT-MQM & 17090 & 549835 &  MQM\\
    \bottomrule
  \end{tabular}
  \end{tabular}
  \caption{Statistics of our used datasets averaged over language pairs. For
  each dataset we report the number of sentences we evaluated on, the amount of
  tokens in these, and the type of human annotation. The number of tokens
  refers to source sentences.}%
  \label{tab:dataset-statistics}
\end{table*}

\end{document}